\documentclass[10pt,twocolumn,letterpaper]{article}

\usepackage[pagenumbers]{cvpr}      %

\usepackage[dvipsnames]{xcolor}

\definecolor{cvprblue}{rgb}{0.21,0.49,0.74}
\usepackage[pagebackref,breaklinks,colorlinks,citecolor=cvprblue]{hyperref}

\DeclareRobustCommand{\mathbf}[1]{\bm{#1}}
\newcommand{\bx}{\mathbf{x}}

\newcommand{\Dbase}{\mathcal{D}_{\text{base}}}
\newcommand{\Dnovel}{\mathcal{D}_{\text{novel}}}

\usepackage{multirow}
\usepackage{multicol}

\title{TransMed: Large Language Models Enhance Vision Transformer for Biomedical Image Classification}
\author{%
    \textbf{Kaipeng Zheng$^1$
    \quad
    Weiran Huang$^{1,}$\thanks{Correspondence to Weiran Huang (weiran.huang@outlook.com).}
    \quad
    Lichao Sun$^2$}\\[0.3cm]
    $^1$ Qing Yuan Research Institute, SEIEE, Shanghai Jiao Tong University\\[0.1cm]
    $^2$ Lehigh University
}

\begin{document}
\maketitle
\begin{abstract}
Few-shot learning has been studied to adapt models to tasks with very few samples.
It holds profound significance, particularly in clinical tasks, due to the high annotation cost of medical images.
Several works have explored few-shot learning on medical images, yet they still require a large number of medical images for pre-training models to gain domain-specific priors.
Vision foundation models recently have achieved remarkable success in natural images. 
Hence, adapting rapidly advancing vision foundation models from natural images to few-shot clinical tasks holds great promise.
MedFMC~\cite{medfmc} has recently organized a challenge to shed more light on this topic at NeurIPS 2023.
In this work, we present our challenge solution.
We observe that a simple variant of fine-tuning with partial freezing shows remarkable performance.
Empirical evidence demonstrates that this approach could outperform various common fine-tuning methods under limited sample sizes. 
Additionally, we explore enhanced utilization of semantic supervision to boost performance. 
We propose a novel approach that contextualizes labels via large language models (LLMs).
Our findings reveal that the context generated by LLMs significantly enhances the discrimination of semantic embeddings for similar categories, resulting in a notable performance improvement of $3\%$-$5\%$ in 1-shot settings compared to commonly employed one-hot labels and other semantic supervision methods.
Our solution secures the 1st place in the MedFMC challenge.
\end{abstract}
 
\section{Introduction}
\label{sec:intro}
Deep learning has yielded significant accomplishments in computer vision, bringing substantial benefits to the advancement of the medical domain.
However, many methods rely on high-quality annotated data, incurring a substantial training cost.
The reality is that annotating medical images requires highly specialized knowledge and consumes considerable time for data collection and labeling. 
Yet, given the dynamic nature of medical tasks in real-world scenarios, it is impractical to invest such high costs in data annotation for every new task.
In such a scenario, using fewer annotations is not only more acceptable but also a practical and reasonable choice. 
Therefore, the approach of few-shot learning, utilizing a minimal number of annotated data samples to adapt models, holds profound significance in the medical domain.
\begin{figure}[t]
\centering
    \includegraphics[width=0.95\linewidth,trim=3cm 0cm 3cm 0cm, clip]{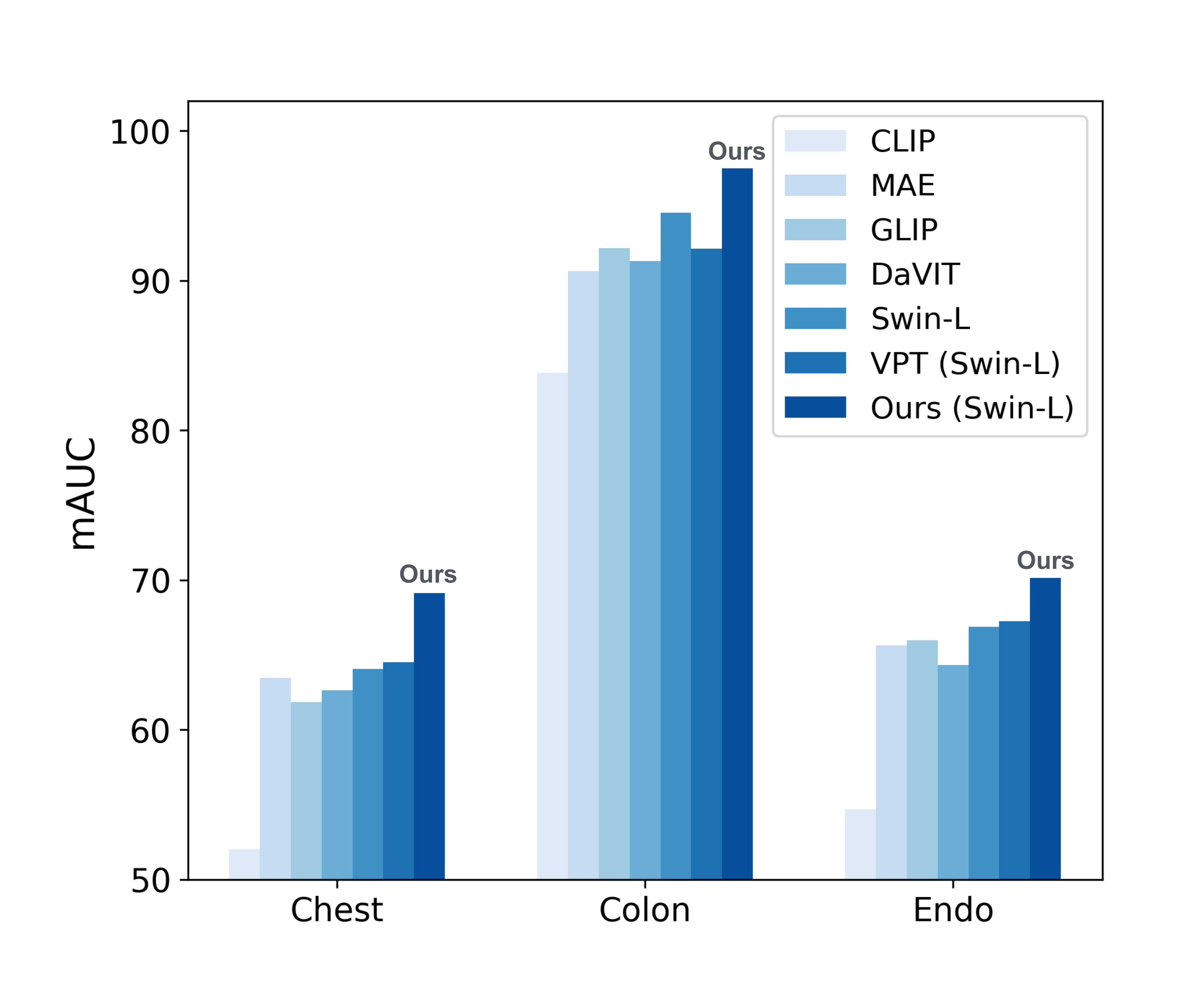}
    \caption{Performance overview on MedFMC in the 5-shot setting. Our approach significantly outperforms the standard adaptation of various foundation models, as well as the state-of-the-art few-shot learning technique, Visual Prompt Tuning \cite{vpt}.
}
\label{fig:intro1} \end{figure}

Training a model from scratch on a few-shot clinical task is a daunting task.
A practical solution is to leverage a pre-trained model for adaptation.
Previous studies \cite{metamed,pfemed,Discriminative-ensemble} have investigated few-shot learning methods for clinical tasks. 
However, the models they employ are pre-trained on medical images.
Collecting the required medical image data for pre-training also demands a significant investment of time.
Recently, numerous vision foundation models \cite{vit,swin,clip,glip,davit,mae} have emerged in natural images, continuously pushing the state-of-the-art performance on ImageNet. 
Hence, one intuitive approach is to directly leverage the vision foundation models in natural images for adaptation to few-shot clinical tasks, without using any external medical images for pretraining.

Prior research \cite{whatmakes,sam-adapter} has also investigated the transfer learning of pre-trained models from natural images to the medical domain. 
However, they still rely on a considerable amount of data for fine-tuning.
In the context of few-shot scenarios, and sometimes in extreme cases where only one data sample is available, this becomes fundamentally different due to the mismatch between the bulky architecture of foundation models and the extremely limited training data. 
Therefore, how to effectively harness the advantages of foundation models from natural images to adapt to few-shot clinical tasks poses a challenging endeavor.
Recently, to shed more light on this issue, MedFMC \cite{medfmc} intentionally constructs a dataset and has organized a challenge at NeurIPS 2023. 
Participants are free to use any foundation model on natural images to adapt the given few-shot clinical tasks, ensuring no medical images are used in pretraining.
The purpose is to uncover the potential of adapting foundation models from natural images to few-shot clinical tasks.

In this paper, we present our solution for the MedFMC challenge, where our approach secures the 1st place.
Given a foundation model pretrained on natural images, due to the substantial domain disparity and extremely limited sample sizes, common adaption methods could suffer severe performance degradation.
We find that this actually can be alleviated with a simple variant of fine-tuning, by freezing several shallow layers of the foundation model and then fine-tuning the remaining layers. 
Despite its simplicity, we verify that it is highly efficient in the adaption of few-shot clinical tasks, yielding high performance across all different settings, even surpassing advanced techniques like LoRA \cite{lora} and Visual Prompt Tuning \cite{vpt}.
Furthermore, we explore how to provide more effective semantic guidance on the adaption of few-shot clinical tasks for further performance gains.
Particularly, our focus is on settings where only category name annotations are available, without detailed medical case reports, since high-quality image-text pairs are not always available in real-world scenarios.
Previous research \cite{CITE} proposes to use the encoding of category names to guide model training.
However, we discover that this could easily lead to high similarities between all lesion types, resulting in inaccurate characterization of inter-class relationships.
To solve this, we propose a novel approach using a large language model for label contextualization, which could successfully increase the separation between the embeddings of different lesion types and achieve significant performance improvements.
Figure \ref{fig:intro1} illustrates the performance improvements achieved by plugging our solution into the swin-transformer. Compared to the baseline method, our approach achieves a $5\%-10\%$ improvement.

Our contributions can be summarized as follows:

\begin{itemize}    
    \item We conduct a thorough analysis of various methods on the adaption of foundation models from natural images to few-shot clinical tasks. We observe that a simple variant of fine-tuning achieves remarkable performance, significantly outperforming standard fine-tuning, as well as advanced techniques like visual prompt tuning and LoRA.

    \item In addition, we explore how to more effectively utilize semantic supervision to further enhance adaptation performance. We propose a novel method that leverages large language models for contextualized labeling, enabling the semantic embeddings of similar diseases to be well-distinguished. This approach achieves significant improvement compared to common practices like one-hot labels or category name encoding.
    
    \item Our final solution secures the championship in the NeurIPS 2023 MedFMC Challenge. Compared to the baseline method of an advanced few-shot learning technique, visual prompt tuning, our approach achieved a 5\%--10\% overall performance improvement.
\end{itemize}
\begin{figure*}[t]
\centering
        \includegraphics[width=0.96\textwidth,trim=0cm 2cm 0cm 0cm, clip]{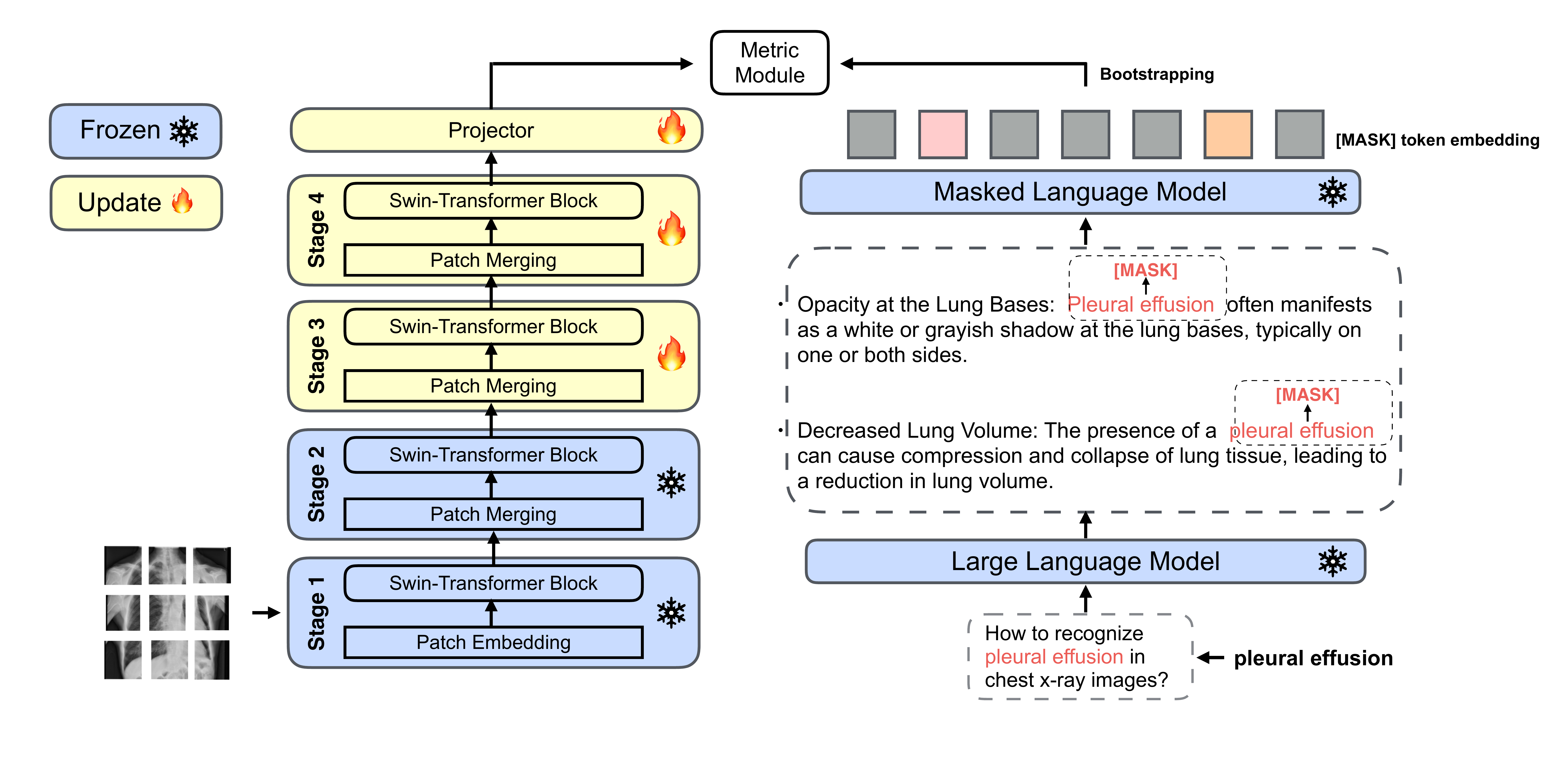}
    \caption{Overall framework of our method. Given an image and a category name, we first utilize a large language model to generate contextual descriptions. Then we replace the category names in the description with [MASK] tokens and extract the embeddings of the [MASK] token using a masked language model as the semantic representation of the category name. We then align the visual embedding of the input image with the textual embedding by fine-tuning the vision foundation models with partial freezing.}
\label{fig:overall} \end{figure*}
\section{Related Work}
\noindent\textbf{Medical Image Classification.}
Deep learning has greatly facilitated research in medical image classification. 
Benefiting from the rapid advancement of vision foundation models on natural images, recent studies \cite{medclip,medclip2,pmc-clip,knowledge-guide} have also explored the use of large-scale data for pretraining models on medical images. 
For instance, MedCLIP \cite{medclip}, MI-Zero \cite{medclip2}, PMC-CLIP \cite{pmc-clip} utilize contrastive learning with a large amount of medical image-text paired data for pretraining, while KAD \cite{knowledge-guide} leverages knowledge graphs to assist pretraining for chest X-ray image diagnosis. 
Although these models may demonstrate high expertise in certain tasks, real-world clinical tasks are highly diverse and often exhibit significant domain variations.
Moreover, the collection of massive medical image data is both labor-intensive and time-consuming. 
Hence, there is a necessity to explore efficient methods for adapting models in scenarios with limited data.

\noindent\textbf{Few-Shot Learning.}
Few-shot learning in natural images has been extensively investigated. 
Meta-learning \cite{maml,proto,relationnet} has been widely studied, with the majority of methods being metric-learning-based methods \cite{proto,relationnet,deepemd}. 
Given a query sample, they determine the semantic similarity between the query sample and the support sample by calculating the similarity between the extracted features of the samples.
Recent research \cite{baseline++,baseline2,pmf} has demonstrated that a standard paradigm of transfer learning, pre-training followed by fine-tuning the last linear layer, can achieve highly competitive performance compared to meta-learning-based methods.
More recently, by introducing continuous prompt tuning in vision Transformers, visual prompt tuning \cite{vpt} has achieved success in fine-tuning vision Transformers \cite{vit,swin} with limited samples.

Previous works \cite{metamed,pfemed} have also explored few-shot learning in medical images, drawing inspiration from meta-learning paradigms on natural images. 
However, this often involves collecting a large number of medical images for pre-training models in advance to acquire domain-specific priors.
To align with the rapid progress of foundation models on natural images, in this work, we investigate directly adapting foundation models from natural images to few-shot clinical tasks.
CITE \cite{CITE} has recently conducted preliminary explorations into this issue.
They employ visual prompt tuning to fine-tune models pre-trained on natural images in a few-shot clinical task, using encoded category names as supervision.
On the contrary, in our study, we provide a comprehensive understanding of this issue on the latest challenging dataset, MedFMC, which encompasses various clinical tasks.
In addition, we propose efficient fine-tuning with partial freezing, as well as LLM-contextualized semantic guidance, which leads to significant performance improvements compared with CITE \cite{CITE} across different modalities of clinical tasks.

\noindent\textbf{Language Supervision for Image Classification.}
Language supervision has recently been extensively studied to guide the pretraining of models in image classification \cite{clip,ram}. 
Compared with common one-hot labels, text often contains descriptions of semantic relationships between categories, providing rich information about inter-class relations.
However, this requires a substantial amount of image-text paired data, which entails significant time and manpower for collection.
Prior research \cite{am3,traml,coop,attribute-proto} has also explored the use of language supervision in the context of few-shot learning or zero-shot learning.
Methods like AM3 \cite{am3} and TRAML \cite{traml} use embeddings of category names to enhance the representation of class prototypes for few-shot learning. 
More recently, CoOp \cite{coop}, VS-Alignment \cite{vs-align} and FILM \cite{film} employ contrastive learning to align visual and textual representations, while \citep{semantic-prompt} utilizes encoded category descriptions to assist in training the visual backbone network.
These approaches rely on individual category names \cite{am3,traml}, existing category descriptions \cite{attribute-proto,vs-align}, or manual and learnable prompts \cite{film,coop} for aligning visual and semantics embedding, without full utilization of contextual information.
In addition, they all investigate natural images, whereas the adaption of clinical tasks remains largely unexplored.
CITE \cite{CITE} recently explores using category names to guide model adaptation in few-shot clinical tasks.
However, we reveal that the simple category name encoding could make the semantic embeddings of different lesions highly similar.
In contrast to previous studies, we propose a novel semantic guidance method by leveraging large language models for contextualized labeling for few-shot clinical tasks where each image is annotated only with category names, which has not been explored before.

\section{Problem Definition}
Typically, a few-shot learning task comprises a support set $S$ and a query set $Q$. 
The goal is to correctly classify the query set using the support set.
The support set is typically composed of an $N$-way, $K$-shot setup, indicating there are $N$ classes in total, each with $K$ samples. $K$ is often a small value, representing a few-shot task.

Training a model from scratch on a few-shot task is extremely challenging due to the severe limitation of the sample size. 
In practice, prior research often focuses on how to adapt a model pretrained on a base dataset $\Dbase$ to a novel few-shot dataset $\Dnovel$, where $\Dnovel$ and $\Dbase$ have no overlap in terms of categories.
Particularly, in this work, we investigate the adaptation of models pretrained on natural images to few-shot clinical tasks. 
Importantly, we ensure that no medical images are involved in the model's pretraining. 
This means that $\Dnovel$ and $\Dbase$ not only have no overlap in terms of categories but also exhibit significant domain differences.
In addition, we particularly explore a challenging scenario where we only have a single disease name annotated for each medical image, unlike previous research \cite{medclip,medclip2,pmc-clip} where detailed clinical reports are available for model training.

\section{
Method}
The overall framework of our approach is illustrated in Figure \ref{fig:overall}. 
To adapt the foundation model from natural images to few-shot clinical tasks, we employ an efficient fine-tuning method that outperforms not only common fine-tuning but also advanced few-shot learning methods like visual prompt tuning. 
Additionally, we introduce a novel approach of leveraging contextualized lesion labels from a large language model for semantic guidance, replacing the commonly used one-hot labels. 
This is verified to be crucial for producing well-differentiated class semantic embeddings, particularly in complex scenarios with numerous categories, hence providing effective semantic guidance for adaptation.
\subsection{Efficient Fine-Tuning with Partial Freezing}
A challenge in adapting foundation models arises from their bulky architecture and extensive parameterization. 
Tuning the parameters of the entire model with just a handful of samples is a daunting task.
Conversely, freezing the feature extractor may result in suboptimal feature extraction, given the substantial domain gap between clinical tasks and natural images.
\begin{table*}[t]
\renewcommand{\arraystretch}{1.1}
\setlength{\tabcolsep}{6pt}
\centering
\resizebox{\textwidth}{!}{%
\begin{tabular}{l|ccc|ccc|ccc}
\toprule
\multirow{2}{*}{\textbf{Method}}
& \multicolumn{3}{c}{\textbf{Chest}}                                                 & \multicolumn{3}{|c}{\textbf{Colon}}                                                & \multicolumn{3}{|c}{\textbf{Endo}}                                                                                         \\
 & 1-shot & 5-shot & 10-shot & 1-shot & 5-shot & 10-shot & 1-shot & 5-shot & 10-shot \\ \midrule
CLIP (ViT-B)                        
& \normalsize{51.52} $\pm$ \text{\footnotesize{1.35}}       & \normalsize{52.03} $\pm$ \text{\footnotesize{0.51}}       & \normalsize{52.34} $\pm$ \text{\footnotesize{1.09}} 
 & \normalsize{78.22} $\pm$ \text{\footnotesize{6.41}} 
  & \normalsize{83.85} $\pm$ \text{\footnotesize{5.86}} 
    & \normalsize{85.53} $\pm$ \text{\footnotesize{1.98}} 
    & \normalsize{48.93} $\pm$ \text{\footnotesize{4.18}}
  & \normalsize{54.69} $\pm$ \text{\footnotesize{5.86}}
    & \normalsize{59.53} $\pm$ \text{\footnotesize{1.55}}
                            \\
MAE (ViT-B)                        
& \normalsize{54.64} $\pm$ \text{\footnotesize{1.59}}       & \normalsize{63.47} $\pm$ \text{\footnotesize{0.60}}       & \normalsize{66.08} $\pm$ \text{\footnotesize{1.41}} 
  & \normalsize{82.50} $\pm$ \text{\footnotesize{4.60}} 
  & \normalsize{90.62} $\pm$ \text{\footnotesize{2.74}} 
    & \normalsize{94.27} $\pm$ \text{\footnotesize{2.22}}
     & \normalsize{58.19} $\pm$ \text{\footnotesize{5.20}}
    & \normalsize{65.66} $\pm$ \text{\footnotesize{2.05}}
     & \normalsize{69.43} $\pm$ \text{\footnotesize{1.01}}
                          \\
Sup ViT-B                            
& \normalsize{55.13} $\pm$ \text{\footnotesize{1.72}}       & \normalsize{63.28} $\pm$ \text{\footnotesize{0.63}} 
 & \normalsize{65.16} $\pm$ \text{\footnotesize{0.56}}
  & \normalsize{84.87} $\pm$ \text{\footnotesize{3.32}} 
& \normalsize{93.26} $\pm$ \text{\footnotesize{2.03}} 
& \normalsize{95.51} $\pm$ \text{\footnotesize{1.25}}
& \normalsize{57.02} $\pm$ \text{\footnotesize{8.75}}
& \normalsize{65.24} $\pm$ \text{\footnotesize{3.24}}
& \normalsize{68.51} $\pm$ \text{\footnotesize{1.38}}                     \\
GLIP (Swin-L)                      
& \normalsize{54.87} $\pm$ \text{\footnotesize{1.49}}   
& \normalsize{61.87} $\pm$ \text{\footnotesize{0.79}}
& \normalsize{64.85} $\pm$ \text{\footnotesize{1.54}}
  & \normalsize{86.41} $\pm$ \text{\footnotesize{3.32}}
& \normalsize{92.18} $\pm$ \text{\footnotesize{1.98}}
& \normalsize{95.57} $\pm$ \text{\footnotesize{1.86}}
& \normalsize{60.17} $\pm$ \text{\footnotesize{4.61}}
& \normalsize{65.99} $\pm$ \text{\footnotesize{3.52}}
& \normalsize{70.69} $\pm$ \text{\footnotesize{0.85}}
                         \\
Sup Swin-L                      
& \normalsize{56.21} $\pm$ \text{\footnotesize{1.67}}  
& \normalsize{64.07} $\pm$ \text{\footnotesize{1.16}} 
& \normalsize{65.94} $\pm$ \text{\footnotesize{0.75}} 
& \normalsize{85.35} $\pm$ \text{\footnotesize{5.27}} 
& \normalsize{94.54} $\pm$ \text{\footnotesize{2.51}} 
& \normalsize{96.66} $\pm$ \text{\footnotesize{1.39}} 
& \normalsize{61.49} $\pm$ \text{\footnotesize{4.65}} 
& \normalsize{66.87} $\pm$ \text{\footnotesize{2.02}} 
& \normalsize{72.39} $\pm$ \text{\footnotesize{0.92}} 
                      \\
DAViT                               
& \normalsize{53.73} $\pm$ \text{\footnotesize{1.35}}       & \normalsize{62.63} $\pm$ \text{\footnotesize{0.85}} 
 & \normalsize{64.80} $\pm$ \text{\footnotesize{0.55}} 
 & \normalsize{85.36} $\pm$ \text{\footnotesize{4.39}} 
  & \normalsize{91.29} $\pm$ \text{\footnotesize{4.08}}
& \normalsize{95.21} $\pm$ \text{\footnotesize{1.72}} 
& \normalsize{56.18} $\pm$ \text{\footnotesize{6.00}} 
& \normalsize{64.34} $\pm$ \text{\footnotesize{5.11}} 
& \normalsize{68.26} $\pm$ \text{\footnotesize{2.73}} 
                         \\ \midrule
CITE \cite{CITE}                                
& \normalsize{58.64} $\pm$ \text{\footnotesize{1.35}}       & \normalsize{65.72} $\pm$ \text{\footnotesize{0.93}} 
 & \normalsize{67.35} $\pm$ \text{\footnotesize{0.98}} 
& \normalsize{83.95} $\pm$ \text{\footnotesize{5.58}} 
& \normalsize{88.90} $\pm$ \text{\footnotesize{1.51}} 
& \normalsize{91.29} $\pm$ \text{\footnotesize{3.43}} 
& \normalsize{59.54} $\pm$ \text{\footnotesize{8.28}} 
& \normalsize{64.55} $\pm$ \text{\footnotesize{5.04}} 
& \normalsize{67.52} $\pm$ \text{\footnotesize{1.88}}  \\
VPT (Swin-L) 
& \normalsize{56.37} $\pm$ \text{\footnotesize{1.15}} 
& \normalsize{64.51} $\pm$ \text{\footnotesize{1.34}} 
& \normalsize{64.29} $\pm$ \text{\footnotesize{1.63}} 
& \normalsize{81.67} $\pm$ \text{\footnotesize{3.80}}
& \normalsize{92.13} $\pm$ \text{\footnotesize{3.55}}
& \normalsize{95.42} $\pm$ \text{\footnotesize{0.56}} 
& \normalsize{60.13} $\pm$ \text{\footnotesize{6.13}} 
& \normalsize{67.26} $\pm$ \text{\footnotesize{2.23}}
& \normalsize{70.99} $\pm$ \text{\footnotesize{1.31}}           \\
Ours (Swin-L)                                
& \textbf{\normalsize{62.95}} $\pm$ \text{\footnotesize{0.19}}  
& \textbf{\normalsize{69.14}} $\pm$ \text{\footnotesize{0.60}} 
& \textbf{\normalsize{71.31}} $\pm$ \text{\footnotesize{0.37}} 
& \textbf{\normalsize{91.50}} $\pm$ \text{\footnotesize{2.47}} 
& \textbf{\normalsize{97.51}} $\pm$ \text{\footnotesize{1.16}} 
& \textbf{\normalsize{97.85}} $\pm$ \text{\footnotesize{1.25}} 
& \textbf{\normalsize{62.51}} $\pm$ \text{\footnotesize{4.65}} 
& \textbf{\normalsize{70.13}} $\pm$ \text{\footnotesize{2.37}} 
& \textbf{\normalsize{74.85}} $\pm$ \text{\footnotesize{0.61}} \\ 

\bottomrule
\end{tabular}}
\caption{Comparison with baselines in 1-shot, 5-shot, and 10-shot settings across all tasks in MedFMC. The average mAUC on the validation set is reported.}
\label{main results}
\end{table*}

\begin{table*}[t]
\renewcommand{\arraystretch}{1.1}
\setlength{\tabcolsep}{6pt}
\centering
\resizebox{\linewidth}{!}{%
\begin{tabular}{c|c|ccc|ccc|ccc}
\toprule
\multirow{2}{*}{\textbf{Backbone}}& \multirow{2}{*}{\textbf{Adaption Method}}& \multicolumn{3}{c}{Chest} & \multicolumn{3}{|c}{Colon}  & \multicolumn{3}{|c}{Endo}                                                                                         \\
 &  & \textbf{1-shot} & \textbf{5-shot} & \textbf{10-shot} & \textbf{1-shot} & \textbf{5-shot} & \textbf{10-shot} & \textbf{1-shot} & \textbf{5-shot} & \textbf{10-shot} \\ \midrule
\multirow{6}{*}{Swin-L}               & Full-Model Fine-Tuning     & \normalsize{58.89} $\pm$ \text{\footnotesize{0.74}}             & \normalsize{65.10} $\pm$ \text{\footnotesize{0.82}}             & \normalsize{67.25} $\pm$ \text{\footnotesize{0.65}}             & \normalsize{87.65} $\pm$ \text{\footnotesize{2.82}}   
& \normalsize{94.21} $\pm$ \text{\footnotesize{3.81}}   
& \normalsize{96.35} $\pm$ \text{\footnotesize{0.95}} 
& \normalsize{59.35} $\pm$ \text{\footnotesize{5.02}}             & \normalsize{68.21} $\pm$ \text{\footnotesize{2.76}}             & \normalsize{69.72} $\pm$ \text{\footnotesize{1.52}}                            \\
& Linear Probe   

& \normalsize{54.18} $\pm$ \text{\footnotesize{1.19}}        & \normalsize{61.16} $\pm$ \text{\footnotesize{0.89}}        & \normalsize{64.30} $\pm$ \text{\footnotesize{0.66}}             & \normalsize{84.36} $\pm$ \text{\footnotesize{5.21}}        & \normalsize{94.50} $\pm$ \text{\footnotesize{1.79}}     & \normalsize{95.67} $\pm$ \text{\footnotesize{0.73}}             
& \normalsize{55.55} $\pm$ \text{\footnotesize{6.53}}             & \normalsize{64.53} $\pm$ \text{\footnotesize{3.72}}             & \normalsize{72.15} $\pm$ \text{\footnotesize{0.67}}                             \\
& Adapter                                                        & \normalsize{56.21} $\pm$ \text{\footnotesize{1.67}}  
& \normalsize{64.07} $\pm$ \text{\footnotesize{1.16}} 
& \normalsize{65.94} $\pm$ \text{\footnotesize{0.75}} 
& \normalsize{85.35} $\pm$ \text{\footnotesize{5.27}} 
& \normalsize{94.54} $\pm$ \text{\footnotesize{2.51}} 
& \normalsize{96.66} $\pm$ \text{\footnotesize{1.39}} 
& \normalsize{61.49} $\pm$ \text{\footnotesize{4.65}} 
& \normalsize{66.87} $\pm$ \text{\footnotesize{2.02}} 
& \normalsize{72.39} $\pm$ \text{\footnotesize{0.92}}                  \\
& LoRA                                                           & \normalsize{53.43} $\pm$ \text{\footnotesize{1.64}}  
& \normalsize{65.54} $\pm$ \text{\footnotesize{0.79}} 
& \normalsize{69.49} $\pm$ \text{\footnotesize{0.13}} 
& \normalsize{87.84} $\pm$ \text{\footnotesize{5.63}} 
& \normalsize{97.21} $\pm$ \text{\footnotesize{1.23}} 
& \normalsize{97.49} $\pm$ \text{\footnotesize{1.80}} 
& \normalsize{58.98} $\pm$ \text{\footnotesize{4.95}} 
& \normalsize{67.69} $\pm$ \text{\footnotesize{3.01}} 
& \normalsize{74.11} $\pm$ \text{\footnotesize{0.53}}                       \\
& Visual Prompt Tuning     
& \normalsize{56.37} $\pm$ \text{\footnotesize{1.15}} 
& \normalsize{64.51} $\pm$ \text{\footnotesize{1.34}} 
& \normalsize{64.29} $\pm$ \text{\footnotesize{1.63}} 
& \normalsize{81.67} $\pm$ \text{\footnotesize{3.80}}
& \normalsize{92.13} $\pm$ \text{\footnotesize{3.55}}
& \normalsize{95.42} $\pm$ \text{\footnotesize{0.56}} 
& \normalsize{60.13} $\pm$ \text{\footnotesize{6.13}} 
& \normalsize{67.26} $\pm$ \text{\footnotesize{2.23}}
& \normalsize{70.99} $\pm$ \text{\footnotesize{1.31}}             \\
& Ours   
& \textbf{\normalsize{59.92}} $\pm$ \text{\footnotesize{1.29}}  
& \textbf{\normalsize{66.90}} $\pm$ \text{\footnotesize{0.62}} 
& \textbf{\normalsize{69.73}} $\pm$ \text{\footnotesize{0.32}} 
& \textbf{\normalsize{91.50}} $\pm$ \text{\footnotesize{2.47}} 
& \textbf{\normalsize{97.51}} $\pm$ \text{\footnotesize{1.16}} 
& \textbf{\normalsize{97.85}} $\pm$ \text{\footnotesize{1.25}} 
& \textbf{\normalsize{62.51}} $\pm$ \text{\footnotesize{4.65}} 
& \textbf{\normalsize{70.13}} $\pm$ \text{\footnotesize{2.37}} 
& \textbf{\normalsize{74.85}} $\pm$ \text{\footnotesize{0.61}}
\\ \midrule
\multirow{6}{*}{ViT-B}               
& Full-Model Fine-Tuning    
& \normalsize{57.06} $\pm$ \text{\footnotesize{1.83}} 
& \normalsize{63.96} $\pm$ \text{\footnotesize{0.20}} 
& \normalsize{66.12} $\pm$ \text{\footnotesize{0.41}} 
& \normalsize{81.55} $\pm$ \text{\footnotesize{3.62}}
& \normalsize{92.74} $\pm$ \text{\footnotesize{3.41}}
& \normalsize{95.23} $\pm$ \text{\footnotesize{0.72}} 
& \normalsize{58.59} $\pm$ \text{\footnotesize{7.44}} 
& \normalsize{64.02} $\pm$ \text{\footnotesize{5.91}}
& \normalsize{70.04} $\pm$ \text{\footnotesize{1.39}}                            \\
& Linear Probe           
& \normalsize{53.69} $\pm$ \text{\footnotesize{0.97}} 
& \normalsize{61.12} $\pm$ \text{\footnotesize{1.60}} 
& \normalsize{65.14} $\pm$ \text{\footnotesize{0.28}} 
& \normalsize{78.88} $\pm$ \text{\footnotesize{7.89}}
& \normalsize{92.72} $\pm$ \text{\footnotesize{1.84}}
& \normalsize{96.03} $\pm$ \text{\footnotesize{1.36}} 
& \normalsize{55.34} $\pm$ \text{\footnotesize{6.83}} 
& \normalsize{63.46} $\pm$ \text{\footnotesize{5.59}}
& \normalsize{70.36} $\pm$ \text{\footnotesize{0.67}}                      \\
& Adapter                
& \normalsize{55.13} $\pm$ \text{\footnotesize{1.72}}       & \normalsize{63.28} $\pm$ \text{\footnotesize{0.63}} 
 & \normalsize{65.16} $\pm$ \text{\footnotesize{0.56}}
  & \normalsize{84.87} $\pm$ \text{\footnotesize{3.32}} 
& \normalsize{93.26} $\pm$ \text{\footnotesize{2.03}} 
& \normalsize{95.51} $\pm$ \text{\footnotesize{1.25}}
& \normalsize{57.02} $\pm$ \text{\footnotesize{8.75}}
& \normalsize{65.24} $\pm$ \text{\footnotesize{3.24}}
& \normalsize{68.51} $\pm$ \text{\footnotesize{1.38}}     \\
& LoRA
& \normalsize{54.84} $\pm$ \text{\footnotesize{1.67}}  
& \normalsize{65.20} $\pm$ \text{\footnotesize{0.55}} 
& \normalsize{67.92} $\pm$ \text{\footnotesize{0.70}} 
& \normalsize{78.19} $\pm$ \text{\footnotesize{6.50}} 
& \normalsize{93.70} $\pm$ \text{\footnotesize{2.38}} 
& \normalsize{94.95} $\pm$ \text{\footnotesize{1.80}} 
& \normalsize{55.91} $\pm$ \text{\footnotesize{9.36}} 
& \normalsize{66.97} $\pm$ \text{\footnotesize{1.72}} 
& \normalsize{72.12} $\pm$ \text{\footnotesize{1.27}}  \\
& Visual Prompt Tuning    
& \normalsize{56.87} $\pm$ \text{\footnotesize{2.00}}  
& \normalsize{63.89} $\pm$ \text{\footnotesize{0.47}} 
& \normalsize{65.60} $\pm$ \text{\footnotesize{0.34}} 
& \normalsize{83.95} $\pm$ \text{\footnotesize{5.58}} 
& \normalsize{88.90} $\pm$ \text{\footnotesize{1.51}} 
& \normalsize{91.29} $\pm$ \text{\footnotesize{3.43}} 
& \normalsize{59.54} $\pm$ \text{\footnotesize{8.28}} 
& \normalsize{64.55} $\pm$ \text{\footnotesize{5.04}} 
& \normalsize{67.52} $\pm$ \text{\footnotesize{1.88}}        \\
& Ours   
& \textbf{\normalsize{59.14}} $\pm$ \text{\footnotesize{1.31}}  
& \textbf{\normalsize{65.47}} $\pm$ \text{\footnotesize{0.61}} 
& \textbf{\normalsize{68.35}} $\pm$ \text{\footnotesize{0.65}} 
& \textbf{\normalsize{87.69}} $\pm$ \text{\footnotesize{3.39}} 
& \textbf{\normalsize{94.54}} $\pm$ \text{\footnotesize{1.79}} 
& \textbf{\normalsize{96.22}} $\pm$ \text{\footnotesize{1.54}} 
& \textbf{\normalsize{61.01}} $\pm$ \text{\footnotesize{6.80}} 
& \textbf{\normalsize{67.62}} $\pm$ \text{\footnotesize{2.13}} 
& \textbf{\normalsize{72.65}} $\pm$ \text{\footnotesize{1.61}}
\\ \bottomrule
\end{tabular}}
\caption{Comparison of different adaption methods on both Swin-Transformer and ViT in 1-shot, 5-shot, and 10-shot settings across all tasks in MedFMC. The average mAUC on the validation set is reported.}
\label{adaption compare}
\end{table*}

In this work, we explore a simple variant of common fine-tuning by selectively freezing parameters within the feature extractor while fine-tuning the remaining ones.
Particularly, given a feature extractor $\mathcal{F} = (f_{\theta_1}, f_{\theta_2}, f_{\theta_3}, ..., f_{\theta_l})$ parameterized by $l$ layers, we find that an extremely simple strategy shows surprisingly high efficiency-just freezing the shallower $N$ layers from $f_{\theta_1}$ to $f_{\theta_N}$ and fine-tuning the remaining deeper layers.
This can be understood as a straightforward compromise to the typical fine-tuning under extreme sample limitations.
We verify that it is generally applicable to commonly used vision transformers like ViT and Swin-Transformer.

\noindent\textbf{Remark.}
Note that the efficiency of fine-tuning with partial freezing actually aligns with findings from previous studies \cite{partial}.
However, it formulates fine-tuning into a complex problem like neural architecture search, by determining the optimal learning rate for each layer within the network, with the goal of finding the optimal subnetwork for adaption.
Although this may find better solutions, we would like to highlight the simplicity of our method, where there is no need for a time-consuming search to determine whether each layer should be frozen.
Empirically, we find that for models like Swin-Transformer, freezing the embedding layer and the transformer layers of the first 2 stages consistently leads to strong performance, significantly surpassing advanced methods such as visual prompt tuning \cite{vpt}.

\subsection{From One-hot Labels to LLM-Contextualized Semantic Guidance}
In classification tasks where only category names are given as the annotation, traditional supervised learning often uses one-hot labels, treating each category as mutually orthogonal. 
However, real-world categories aren't always strictly independent. Some share semantic closeness, like cats and dogs, while others have distinct differences, like cats and ships.
This also holds true in medical tasks. Certain diseases share pivotal features, forming close recognition patterns, while others have distinctive features. 
Using one-hot labels ignores these semantic relationships, making few-shot learning more complex by forcing unnatural orthogonal relationships in category embeddings.

Language supervision has been explored to address this issue, particularly in the context of few-shot learning \cite{am3,traml,CITE}. 
Compared to image annotation data that requires specialized knowledge, a general pre-trained medical language model on PubMed is easily accessible. 
A straightforward approach involves encoding category names as semantic supervision, a method that has been investigated in prior works \cite{CITE}.
However, we highlight that simply encoding category names would result in highly similar embeddings for different classes (see Section \ref{context-matter}).
This leads to an inaccurate description of relationships between categories, causing a significant decline in performance.
The reason arises from the fact that diagnosing lesions involves a specific clinical task, where different lesions exhibit semantic proximity, particularly for a general medical language model. 
This makes their discrimination a fine-grained classification task.
Hence, there is a need for further exploration of how to provide effective semantic guidance on the adaptation of few-shot clinical tasks.

In this work, we propose a novel approach for semantic guidance.
We utilize a Large Language Model (LLM) to contextually expand a single disease name label, introducing richer information about disease diagnosis to differentiate it from other diseases. 
Without loss of generality, take the Chest task as an example. 
Given a class named [CLASS], we leverage an LLM for contextualizing labels, with a very simple query: ``How to recognize [CLASS] in chest X-ray images?''
The LLM then provides answers about key regions in chest X-ray images for diagnosing the disease and the typical features of these regions.
To validate the effectiveness of the answers, we replace the category names in the answers with [MASK] tokens. 
Utilizing a commonly available BERT pretrained on PubMed, we find that in fact correct predictions for the [MASK] token can be made within top-5 candidates in most cases, given the context generated by the LLM.
This implies that the generated contexts indeed provide effective contextual information for the diagnosis of diseases.
We subsequently derive semantic embeddings for category $c$ by extracting features at the [MASK] token $\mathcal{T}_c=(t^c_1, t^c_2, ..., t^c_m)$, where $m$ represents the number of [MASK] tokens in the generated context. 
The multi-label classification likelihood is computed by independently aligning each semantic embedding $\mathcal{T}_c$ with the visual embedding $f_\theta(x_i)$ extracted by the foundation model, as follows,
\begin{align}
P (y=c|\bx)=\frac{1}{1+\exp(-\sum_{i=0}^{m_0} \text{sim} (f_\theta (\bx), t_i^c)\cdot \tau)}, \label{eq1}
\end{align}
where $\text{sim}(\cdot)$ denotes a similarity metric, $m_0$ represents the number of chosen [MASK] tokens, and $\tau$ is a scaling factor.

In training, for each class, a subset $T'_c=(t^c_1,t^c_2,...,t^c_{m_0})$ is bootstrapped from $T_c$ in each epoch to align with visual features using Equation \eqref{eq1}. 
The Binary Cross Entropy (BCE) Loss is employed as the loss function. 
During testing, predictions are made using all [MASK] tokens.

\section{Experiments}
\subsection{MedFM Challenge} 
\begin{figure*}[t]

\centering
        \includegraphics[width=0.88\textwidth,trim=0cm 5cm 0cm 0cm, clip]{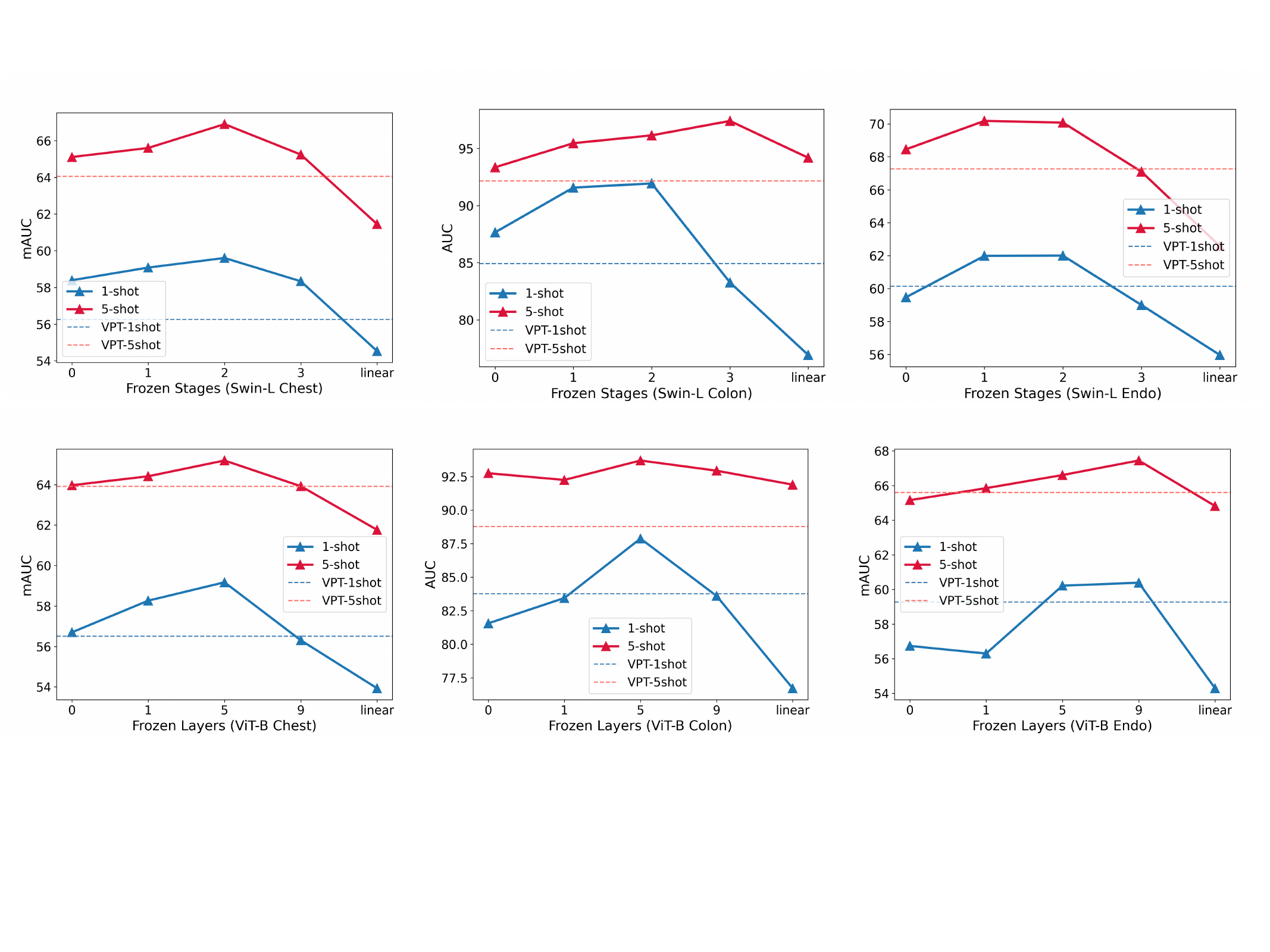}
    \caption{Emperical analysis of setting different numbers of frozen layers. We conduct evaluation on both ViT and Swin-Transformer under 1-shot and 5-shot settings across all datasets. Fine-tuning with partial freezing demonstrates a good compromise between full-model fine-tuning and linear probing, surpassing Visual Prompt Tuning (VPT).}
\label{fig:stage-ablations} \end{figure*}

\begin{figure}[t]
\centering
        \includegraphics[width=0.5\textwidth,trim=1cm 10cm 0cm 2cm, clip]{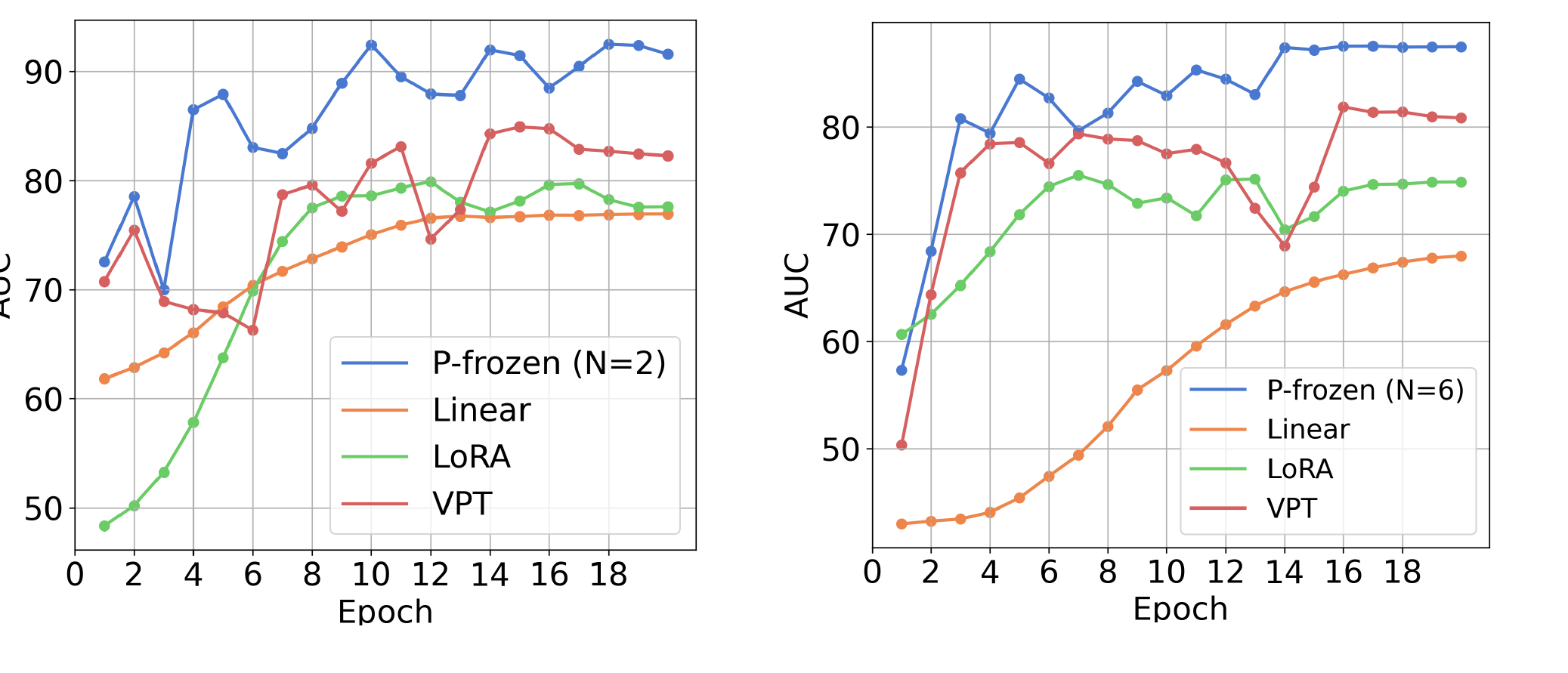}
    \caption{Performance changes on Colon 1-shot tasks as adaptation progresses. Fine-tuning with partial freezing demonstrated superior performance compared to other adaption techniques.}
\label{fig:performance-vurve} \end{figure}

MedFM \cite{medfmc} hosts a challenge at NeurIPS 2023, targeting foundation model adaption on few-shot clinical tasks.
It offers the most recent and comprehensive benchmark for assessing the performance of few-shot learning on medical images.
Note that, unlike previous benchmarks \cite{dataset1,dataset2}, MedFMC is specifically designed to evaluate the adaptation of foundational models from natural images to medical images in few-shot scenarios.
This means that, in the challenge, medical images are not allowed to participate in the pre-training of the models.
Foundation models pre-trained on natural images can be utilized.
In this work, we investigate the three publicly open tasks from the challenge:

\noindent\textbf{ChestDR.} The ChestDR task involves the diagnosis of chest X-ray images, encompassing 19 diseases. 
Each chest X-ray image is multi-labeled. 
Hence, the ChestDR task can be actually interpreted as a 19-way few-shot multi-label classification problem.
The training, validation, and test sets consist of 2140, 2708, and 2626 images, respectively.

\noindent\textbf{ColonPath.} The ColonPath task is a few-shot pathological image diagnosis task, constituting a binary classification problem to discern whether an image depicts a tumor. 
The training, validation, and test sets comprise 5654, 4355, and 10009 images, respectively.

\noindent\textbf{Endo.} The Endo task involves diagnosing colonoscopy images, addressing three abnormalities and tumors, with each image being multi-labeled. 
The training, validation, and test sets consist of 1810, 2055, and 2199 images, respectively.

In the MedFM Challenge, the support sets are given as the fixed subset of the training set based on the number of patients, with 1-shot, 5-shot, and 10-shot scenarios.
The entire validation and test sets are fixed as the query set for performance evaluation.
Since labels for the test set data are not publicly available, we use the validation set as the query set in this work.
All annotations on the three datasets only include class names.
The mean AUC (mAUC) is used as the evaluation metric.

\noindent\textbf{Baseline.}
Adapting Swin-Transformer \cite{swin} with Visual Prompt Tuning \cite{vpt} serves as the official baseline for the MedFMC competition. 
In our comparison, we also include the latest approach, CITE \cite{CITE}, which first investigates the adaptation of foundation models to few-shot clinical tasks.

\subsection{Implementations}
For the vision foundation model, we employ the Swin-Transformer-Large \cite{swin} pretrained on ImageNet21k. 
The parameters in the Swin-Transformer could be divided into four stages based on downsampling. 
We simply set the number of frozen stages to 2, indicating that during fine-tuning, the parameters in the first two stages are partially frozen.
For the semantic guidance, we utilize GPT-4 for label contextualization and BERT pretrained on PubMed as the masked language model to generate semantic embedding. 
We use cosine similarity as the $\text{sim}(\cdot)$ in Eq. \eqref{eq1}, with $\tau$ set to 10.
For the adaption, we employed AdamW \cite{adam} as the optimizer, set the batch size to 4, learning rate to $10^{-4}$. 
The number of adaptation epochs is set to 20.
Data augmentation techniques used in our method are simply center cropping, random cropping, and random horizontal flipping.
\subsection{Results}
Table \ref{main results} illustrates the performance of our approach. 
In comparison with Visual Prompt Tuning \cite{vpt}, a state-of-the-art adaptation technique for Vision Transformers, our method achieves a performance improvement ranging from $2\%$ to $10\%$ across all tasks in 1-shot, 5-shot, and 10-shot settings.
Moreover, our approach outperforms the latest method CITE \cite{CITE} by $3\%$ to $7\%$ across various settings. 
\begin{figure*}[t]
\centering
        \includegraphics[width=0.75\textwidth]{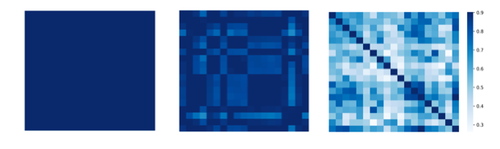}
    \caption{The inter-class correlation matrices obtained by different language supervision methods on the ChestDR task. Left: Encoding category names. Middle: a template method using masked language models without contextualization. Right: our method leveraging large language models for label contextualization. It can be observed that the context generated by large language models plays a crucial role in fine-grained category distinguishing.}
\label{fig:semantic-correlation} \end{figure*}
Intuitively, we also make comparisons with the direct adaption of diverse state-of-the-art foundation models on natural images with varying pre-training datasets (CLIP \cite{clip}, GLIP \cite{glip}, ImageNet-21K), pre-training methods (Supervised (Sup), Self-supervised (MAE) \cite{mae}), and model architectures (ViT-B \cite{vit}, Swin-Transformer-L \cite{swin}, DaViT \cite{davit}) using an adapter-based method \cite{adapter}.
It can be observed that advanced foundation models trained on natural images do not exhibit superiority in adaptation to few-shot clinical tasks. 
This observation actually aligns with previous findings \cite{close-look-again}, indicating that performance on downstream tasks is nearly uncorrelated to the model's performance on ImageNet when the downstream task involves significant domain differences.
\subsection{Fine-Tuning with Partial Freezing is Efficient in Few-shot Scenarios}
\noindent\textbf{Performance Improvements.}
We then take a close look at the effect of fine-tuning with partial freezing in few-shot scenarios. 
To demonstrate the effectiveness, we make comparisons with several representative adaptation algorithms, including full model fine-tuning, linear probe, 
LoRA \cite{lora}, and Visual Prompt Tuning \cite{vpt}. 
We conduct evaluations on both Swin-Transformer and ViT.
The results are shown in Table \ref{adaption compare}. 
It can be observed that fine-tuning with partial freezing showcases superior performance, outperforming visual prompt tuning by $2\%$ to $10\%$.

\noindent\textbf{Impact of Frozen Layers.}
We then delve into a more detailed analysis of partial fine-tuning, specifically focusing on the number of frozen layers. 
For a fair comparison, we only varied the number of shallower layers frozen in the model, while keeping all other settings constant. 
Without loss of generality, we conduct evaluations on both Swin-Transformer and ViT.
The experimental results are shown in Figure 4, where the `Frozen Stages' or `Frozen Layers' equal to 0 implies the typical full-model fine-tuning scenario, where no layers' parameters are frozen, and linear' denotes freezing all parameters in the feature extractor, tuning only the parameters of the final linear layer.
It can be observed that, compared to the common full-model fine-tuning and fine-tuning only the last layer, partial freezing generally contributes to a performance improvement of approximately $2\%$ in both 1-shot and 5-shot settings across all different tasks, vastly surpassing the advanced few-shot learning technique, Visual Prompt Tuning.
This reflects that fine-tuning with partial freezing of shallow network parameters is demonstrated to be a well-balanced approach in scenarios with significant domain differences and limited sample sizes, offering a compromise between frozen and fully-tuned models.
In addition, another key observation is that completely freezing the feature extractor leads to a significant performance decline in the majority of cases. 
This contradicts previous findings in several few-shot learning works on natural images, where the feature extractor is typically frozen during adaption \cite{baseline++,baseline2,pmf}. 
This discrepancy can be attributed to the significant domain differences between medical and natural images.

In Figure \ref{fig:performance-vurve}, we further depict the performance changes of different adaption methods on the validation set as the adaptation progresses. 
It can be observed that the improvements achieved by fine-tuning with partial freezing are evident compared with existing adaption techniques in few-shot learning.

\subsection{Context Matters for Effective Semantic Guidance} \label{context-matter}
Next, we analyze the effect of different language supervision methods on the adaption of few-shot clinical tasks, in cases where only category names are annotated.
\begin{table*}[]
\renewcommand{\arraystretch}{1}
\setlength{\tabcolsep}{6pt}
\centering
\begin{tabular}{lccccc}
\toprule
\multicolumn{1}{c}{\textbf{Method}} & \textbf{Adaption}              & \textbf{Semantic Supervision} & \textbf{1-shot} & \textbf{5-shot} & \textbf{10-shot} \\ \midrule
Swin-L                              & fine-tuning (frozen-stage = 1) & None                
& \normalsize{59.92} $\pm$ \text{\footnotesize{1.29}}  
& \normalsize{66.90} $\pm$ \text{\footnotesize{0.62}} 
& \normalsize{69.73} $\pm$ \text{\footnotesize{0.32}}       \\
Swin-L                              & fine-tuning (frozen-stage = 1) & class-name         
& \normalsize{59.08} $\pm$ \text{\footnotesize{1.22}}     & \normalsize{68.02} $\pm$ \text{\footnotesize{0.66}} 
 & \normalsize{70.37} $\pm$ \text{\footnotesize{0.78}}        \\
Swin-L                              & fine-tuning (frozen-stage = 1) & template        
& \normalsize{57.32} $\pm$ \text{\footnotesize{4.59}}     & \normalsize{68.38} $\pm$ \text{\footnotesize{0.50}} 
 & \normalsize{70.72} $\pm$ \text{\footnotesize{0.63}}         \\
Swin-L                              & fine-tuning (frozen-stage = 1) & context (ours)           & 
 \textbf{\normalsize{62.95}} $\pm$ \text{\footnotesize{0.19}}  
& \textbf{\normalsize{69.24}} $\pm$ \text{\footnotesize{0.60}} 
& \textbf{\normalsize{71.41}} $\pm$ \text{\footnotesize{0.37}} 
\\ \midrule
ViT-B/16                            & visual prompt tuning           & None                          
& \normalsize{56.87} $\pm$ \text{\footnotesize{2.00}}  
& \normalsize{63.89} $\pm$ \text{\footnotesize{0.47}} 
& \normalsize{65.60} $\pm$ \text{\footnotesize{0.34}}             \\
ViT-B/16                            & visual prompt tuning           & class-name                    
& \normalsize{58.64} $\pm$ \text{\footnotesize{1.35}}     & \normalsize{65.72} $\pm$ \text{\footnotesize{0.93}} 
 & \normalsize{67.35} $\pm$ \text{\footnotesize{0.98}} 
 \\
ViT-B/16                            & visual prompt tuning           & template                   
& \normalsize{54.51} $\pm$ \text{\footnotesize{3.30}}     & \normalsize{65.94} $\pm$ \text{\footnotesize{0.70}} 
 & \normalsize{67.96} $\pm$ \text{\footnotesize{0.51}}         \\
ViT-B/16                            & visual prompt tuning           & context (ours)         
& \textbf{\normalsize{59.66}} $\pm$ \text{\footnotesize{1.69}}     & \textbf{\normalsize{66.31}} $\pm$ \text{\footnotesize{0.94}} 
 & \textbf{\normalsize{68.58}} $\pm$ \text{\footnotesize{0.40}}  \\ \bottomrule
\end{tabular}
\caption{Comparison of different language supervision methods on the ChestDR task in 1-shot, 5-shot, and 10-shot settings, where ``None'' indicates the use of only one-hot labels. The average mAUC on the validation set is reported.}
\label{semantic compare}
\end{table*}

\noindent\textbf{LLM Provides Effective Contexts.}
To validate the effectiveness of label contextualizing with large language models, we compare our approach with recent language supervision methods in few-shot learning, as follows:
Direct encoding of category names, as used in several previous few-shot learning techniques \cite{am3,traml} and CITE \cite{CITE}.
A template-based approach using masked language models to generate category embedding, which is proposed in the latest research \cite{film} in few-shot learning:
Given a category name [CLASS], it constructs a template like: ``The symptom of [CLASS] in chest x-ray image is [MASK].'' The embedding of the [MASK] token is then used as the semantic embedding for the category.
Given the complexity of the Chest task, which involves a classification of 19 categories of chest X-ray diagnoses, our analysis mainly focuses on the Chest task.
Figure \ref{fig:semantic-correlation} illustrates the inter-class correlation matrix of the semantic embedding for the 19 chest X-ray diseases, generated by different language supervision methods.
It can be observed that both class name encoding and the template-based method result in extremely high inter-class similarity, making it challenging for the model to distinguish between different categories.
This difficulty is inherent in clinical tasks where discerning between lesions requires a high level of expertise.
On the other hand, our method effectively widens the distance between categories.
This reflects that the contextualized category names effectively provide fine-grained discriminative information among similar lesions within the same clinical task.

\noindent\textbf{Performance Improvements.}
We then make performance comparisons, evaluating our approach on both Swin-Transformer and ViT through partial fine-tuning and Visual Prompt Tuning. 
The results are presented in Table \ref{semantic compare}.
Our approach consistently outperforms other language supervision methods. 
Notably, when compared to the one-hot labels and other language supervision methods, our method demonstrates a significant improvement of $3\%$-$5\%$ in 1-shot settings. 
Furthermore, the improvements achieved by combining our method with different adaption techniques and different foundation models also highlight its utility.
\section{Conclusion}
In this work, we study the adaptation of foundation models from natural images to few-shot clinical tasks, ensuring no medical data is used in the pre-training.
Our solution secures 1st place in the MedFM challenge at NeurIPS 2023.
We explore a simple fine-tuning variant with partial freezing, which showcases effectiveness in tasks with limited samples and large domain differences, outperforming state-of-the-art adaption methods like Visual Prompt Tuning.
Furthermore, we study how to leverage semantic guidance for further improvements when only category names are given.
We propose to utilize large language models for contextualized labeling, surpassing one-hot labels and category name encoding used in previous studies.
Compared with the baseline, our method demonstrates a $2\%$-$10\%$ performance improvement across all settings in MedFMC.

\clearpage
{
    \small
    \bibliographystyle{ieeenat_fullname}
    \bibliography{main}

\begin{thebibliography}{40}
\providecommand{\natexlab}[1]{#1}
\providecommand{\url}[1]{\texttt{#1}}
\expandafter\ifx\csname urlstyle\endcsname\relax
  \providecommand{\doi}[1]{doi: #1}\else
  \providecommand{\doi}{doi: \begingroup \urlstyle{rm}\Url}\fi

\bibitem[Afham and Rodrigo(2022)]{vs-align}
Mohamed Afham and Ranga Rodrigo.
\newblock Visual-semantic contrastive alignment for few-shot image classification.
\newblock \emph{CoRR}, abs/2210.11000, 2022.

\bibitem[Chen et~al.(2019)Chen, Liu, Kira, Wang, and Huang]{baseline++}
Wei{-}Yu Chen, Yen{-}Cheng Liu, Zsolt Kira, Yu{-}Chiang~Frank Wang, and Jia{-}Bin Huang.
\newblock A closer look at few-shot classification.
\newblock In \emph{ICLR}, 2019.

\bibitem[Chen et~al.(2023)Chen, Si, Zhang, Wang, Wang, and Tan]{semantic-prompt}
Wentao Chen, Chenyang Si, Zhang Zhang, Liang Wang, Zilei Wang, and Tieniu Tan.
\newblock Semantic prompt for few-shot image recognition.
\newblock \emph{CoRR}, abs/2303.14123, 2023.

\bibitem[Dai et~al.(2023)Dai, Yi, Yan, Xu, Hu, Zhang, Li, and Wang]{pfemed}
Zhiyong Dai, Jianjun Yi, Lei Yan, Qingwen Xu, Liang Hu, Qi Zhang, Jiahui Li, and Guoqiang Wang.
\newblock Pfemed: Few-shot medical image classification using prior guided feature enhancement.
\newblock \emph{Pattern Recognit.}, 2023.

\bibitem[Dhillon et~al.(2020)Dhillon, Chaudhari, Ravichandran, and Soatto]{baseline2}
Guneet~Singh Dhillon, Pratik Chaudhari, Avinash Ravichandran, and Stefano Soatto.
\newblock A baseline for few-shot image classification.
\newblock In \emph{8th International Conference on Learning Representations, {ICLR} 2020, Addis Ababa, Ethiopia, April 26-30, 2020}. OpenReview.net, 2020.

\bibitem[Ding et~al.(2022)Ding, Xiao, Codella, Luo, Wang, and Yuan]{davit}
Mingyu Ding, Bin Xiao, Noel Codella, Ping Luo, Jingdong Wang, and Lu Yuan.
\newblock Davit: Dual attention vision transformers.
\newblock In \emph{ECCV}, 2022.

\bibitem[Dosovitskiy et~al.(2021)Dosovitskiy, Beyer, Kolesnikov, Weissenborn, Zhai, Unterthiner, Dehghani, Minderer, Heigold, Gelly, Uszkoreit, and Houlsby]{vit}
Alexey Dosovitskiy, Lucas Beyer, Alexander Kolesnikov, Dirk Weissenborn, Xiaohua Zhai, Thomas Unterthiner, Mostafa Dehghani, Matthias Minderer, Georg Heigold, Sylvain Gelly, Jakob Uszkoreit, and Neil Houlsby.
\newblock An image is worth 16x16 words: Transformers for image recognition at scale.
\newblock In \emph{ICLR}, 2021.

\bibitem[Finn et~al.(2017)Finn, Abbeel, and Levine]{maml}
Chelsea Finn, Pieter Abbeel, and Sergey Levine.
\newblock Model-agnostic meta-learning for fast adaptation of deep networks.
\newblock In \emph{ICML}, 2017.

\bibitem[He et~al.(2022)He, Chen, Xie, Li, Doll{\'{a}}r, and Girshick]{mae}
Kaiming He, Xinlei Chen, Saining Xie, Yanghao Li, Piotr Doll{\'{a}}r, and Ross~B. Girshick.
\newblock Masked autoencoders are scalable vision learners.
\newblock In \emph{CVPR}, 2022.

\bibitem[Hu et~al.(2022{\natexlab{a}})Hu, Shen, Wallis, Allen{-}Zhu, Li, Wang, Wang, and Chen]{lora}
Edward~J. Hu, Yelong Shen, Phillip Wallis, Zeyuan Allen{-}Zhu, Yuanzhi Li, Shean Wang, Lu Wang, and Weizhu Chen.
\newblock Lora: Low-rank adaptation of large language models.
\newblock In \emph{ICLR}, 2022{\natexlab{a}}.

\bibitem[Hu et~al.(2022{\natexlab{b}})Hu, Li, St{\"{u}}hmer, Kim, and Hospedales]{pmf}
Shell~Xu Hu, Da Li, Jan St{\"{u}}hmer, Minyoung Kim, and Timothy~M. Hospedales.
\newblock Pushing the limits of simple pipelines for few-shot learning: External data and fine-tuning make a difference.
\newblock In \emph{CVPR}, 2022{\natexlab{b}}.

\bibitem[Jia et~al.(2022)Jia, Tang, Chen, Cardie, Belongie, Hariharan, and Lim]{vpt}
Menglin Jia, Luming Tang, Bor{-}Chun Chen, Claire Cardie, Serge~J. Belongie, Bharath Hariharan, and Ser{-}Nam Lim.
\newblock Visual prompt tuning.
\newblock In \emph{ECCV}, 2022.

\bibitem[Jiang et~al.(2023)Jiang, Dang, Pang, Zhang, and Huang]{film}
Zihao Jiang, Yunkai Dang, Dong Pang, Huishuai Zhang, and Weiran Huang.
\newblock {FILM:} how can few-shot image classification benefit from pre-trained language models?
\newblock \emph{CoRR}, abs/2307.04114, 2023.

\bibitem[Kingma and Ba(2015)]{adam}
Diederik~P. Kingma and Jimmy Ba.
\newblock Adam: {A} method for stochastic optimization.
\newblock In \emph{ICLR}, 2015.

\bibitem[Li et~al.(2020)Li, Huang, Lan, Feng, Li, and Wang]{traml}
Aoxue Li, Weiran Huang, Xu Lan, Jiashi Feng, Zhenguo Li, and Liwei Wang.
\newblock Boosting few-shot learning with adaptive margin loss.
\newblock In \emph{CVPR}, 2020.

\bibitem[Li et~al.(2022)Li, Zhang, Zhang, Yang, Li, Zhong, Wang, Yuan, Zhang, Hwang, Chang, and Gao]{glip}
Liunian~Harold Li, Pengchuan Zhang, Haotian Zhang, Jianwei Yang, Chunyuan Li, Yiwu Zhong, Lijuan Wang, Lu Yuan, Lei Zhang, Jenq{-}Neng Hwang, Kai{-}Wei Chang, and Jianfeng Gao.
\newblock Grounded language-image pre-training.
\newblock In \emph{CVPR}, pages 10955--10965. {IEEE}, 2022.

\bibitem[Lin et~al.(2023)Lin, Zhao, Zhang, Wu, Zhang, Wang, and Xie]{pmc-clip}
Weixiong Lin, Ziheng Zhao, Xiaoman Zhang, Chaoyi Wu, Ya Zhang, Yanfeng Wang, and Weidi Xie.
\newblock {PMC-CLIP:} contrastive language-image pre-training using biomedical documents.
\newblock In \emph{MICCAI}, 2023.

\bibitem[Liu et~al.(2021)Liu, Lin, Cao, Hu, Wei, Zhang, Lin, and Guo]{swin}
Ze Liu, Yutong Lin, Yue Cao, Han Hu, Yixuan Wei, Zheng Zhang, Stephen Lin, and Baining Guo.
\newblock Swin transformer: Hierarchical vision transformer using shifted windows.
\newblock In \emph{ICCV}, 2021.

\bibitem[Lu et~al.(2023)Lu, Chen, Zhang, Williamson, Chen, Ding, Le, Chuang, and Mahmood]{medclip2}
Ming~Y. Lu, Bowen Chen, Andrew Zhang, Drew F.~K. Williamson, Richard~J. Chen, Tong Ding, Long~Phi Le, Yung{-}Sung Chuang, and Faisal Mahmood.
\newblock Visual language pretrained multiple instance zero-shot transfer for histopathology images.
\newblock In \emph{CVPR}, 2023.

\bibitem[Luo et~al.(2023)Luo, Wu, Zhang, Gao, Xu, and Song]{close-look-again}
Xu Luo, Hao Wu, Ji Zhang, Lianli Gao, Jing Xu, and Jingkuan Song.
\newblock A closer look at few-shot classification again.
\newblock In \emph{ICML}, 2023.

\bibitem[Matsoukas et~al.(2022)Matsoukas, Haslum, Sorkhei, S{\"{o}}derberg, and Smith]{whatmakes}
Christos Matsoukas, Johan~Fredin Haslum, Moein Sorkhei, Magnus S{\"{o}}derberg, and Kevin Smith.
\newblock What makes transfer learning work for medical images: Feature reuse {\&} other factors.
\newblock In \emph{CVPR}, 2022.

\bibitem[Paul et~al.(2021)Paul, Tang, Shen, and Summers]{Discriminative-ensemble}
Angshuman Paul, Yuxing Tang, Thomas~C. Shen, and Ronald~M. Summers.
\newblock Discriminative ensemble learning for few-shot chest x-ray diagnosis.
\newblock \emph{Medical Image Anal.}, 2021.

\bibitem[Radford et~al.(2021)Radford, Kim, Hallacy, Ramesh, Goh, Agarwal, Sastry, Askell, Mishkin, Clark, Krueger, and Sutskever]{clip}
Alec Radford, Jong~Wook Kim, Chris Hallacy, Aditya Ramesh, Gabriel Goh, Sandhini Agarwal, Girish Sastry, Amanda Askell, Pamela Mishkin, Jack Clark, Gretchen Krueger, and Ilya Sutskever.
\newblock Learning transferable visual models from natural language supervision.
\newblock In \emph{ICML}, 2021.

\bibitem[R{\"{u}}ckl{\'{e}} et~al.(2021)R{\"{u}}ckl{\'{e}}, Geigle, Glockner, Beck, Pfeiffer, Reimers, and Gurevych]{adapter}
Andreas R{\"{u}}ckl{\'{e}}, Gregor Geigle, Max Glockner, Tilman Beck, Jonas Pfeiffer, Nils Reimers, and Iryna Gurevych.
\newblock Adapterdrop: On the efficiency of adapters in transformers.
\newblock In \emph{EMNLP}, 2021.

\bibitem[Shakeri et~al.(2022)Shakeri, Boudiaf, Mohammadi, Sheth, Havaei, Ayed, and Kahou]{dataset2}
Fereshteh Shakeri, Malik Boudiaf, Sina Mohammadi, Ivaxi Sheth, Mohammad Havaei, Ismail~Ben Ayed, and Samira~Ebrahimi Kahou.
\newblock {FHIST:} {A} benchmark for few-shot classification of histological images.
\newblock \emph{CoRR}, abs/2206.00092, 2022.

\bibitem[Shen et~al.(2021)Shen, Liu, Qin, Savvides, and Cheng]{partial}
Zhiqiang Shen, Zechun Liu, Jie Qin, Marios Savvides, and Kwang{-}Ting Cheng.
\newblock Partial is better than all: Revisiting fine-tuning strategy for few-shot learning.
\newblock In \emph{AAAI}, 2021.

\bibitem[Singh et~al.(2021)Singh, Bharti, Purohit, Kumar, Singh, and Singh]{metamed}
Rishav Singh, Vandana Bharti, Vishal Purohit, Abhinav Kumar, Amit~Kumar Singh, and Sanjay~Kumar Singh.
\newblock Metamed: Few-shot medical image classification using gradient-based meta-learning.
\newblock \emph{Pattern Recognit.}, 2021.

\bibitem[Snell et~al.()Snell, Swersky, and Zemel]{proto}
Jake Snell, Kevin Swersky, and Richard~S. Zemel.
\newblock Prototypical networks for few-shot learning.
\newblock In \emph{NeurIPSUSA}.

\bibitem[Sun et~al.(2022)Sun, Li, Ding, Huang, Chen, Wang, Yu, and Paisley]{dataset1}
Liyan Sun, Chenxin Li, Xinghao Ding, Yue Huang, Zhong Chen, Guisheng Wang, Yizhou Yu, and John~W. Paisley.
\newblock Few-shot medical image segmentation using a global correlation network with discriminative embedding.
\newblock \emph{Comput. Biol. Medicine}, 140:\penalty0 105067, 2022.

\bibitem[Sung et~al.(2018)Sung, Yang, Zhang, Xiang, Torr, and Hospedales]{relationnet}
Flood Sung, Yongxin Yang, Li Zhang, Tao Xiang, Philip H.~S. Torr, and Timothy~M. Hospedales.
\newblock Learning to compare: Relation network for few-shot learning.
\newblock In \emph{CVPR}, 2018.

\bibitem[Wang et~al.(2023)Wang, Wang, Wang, Li, Da, Liu, Gao, Shen, He, Shen, Duan, Zhao, Li, Qiao, and Zhang]{medfmc}
Dequan Wang, Xiaosong Wang, Lilong Wang, Mengzhang Li, Qian Da, Xiaoqiang Liu, Xiangyu Gao, Jun Shen, Junjun He, Tian Shen, Qi Duan, Jie Zhao, Kang Li, Yu Qiao, and Shaoting Zhang.
\newblock Medfmc: {A} real-world dataset and benchmark for foundation model adaptation in medical image classification.
\newblock \emph{CoRR}, abs/2306.09579, 2023.

\bibitem[Wang et~al.(2022)Wang, Wu, Agarwal, and Sun]{medclip}
Zifeng Wang, Zhenbang Wu, Dinesh Agarwal, and Jimeng Sun.
\newblock Medclip: Contrastive learning from unpaired medical images and text.
\newblock In \emph{EMNLP}. Association for Computational Linguistics, 2022.

\bibitem[Wu et~al.(2023)Wu, Fu, Fang, Liu, Wang, Xu, Jin, and Arbel]{sam-adapter}
Junde Wu, Rao Fu, Huihui Fang, Yuanpei Liu, Zhaowei Wang, Yanwu Xu, Yueming Jin, and Tal Arbel.
\newblock Medical {SAM} adapter: Adapting segment anything model for medical image segmentation.
\newblock \emph{CoRR}, 2023.

\bibitem[Xing et~al.(2019)Xing, Rostamzadeh, Oreshkin, and Pinheiro]{am3}
Chen Xing, Negar Rostamzadeh, Boris~N. Oreshkin, and Pedro~O. Pinheiro.
\newblock Adaptive cross-modal few-shot learning.
\newblock In \emph{NeurIPS}, 2019.

\bibitem[Xu et~al.(2022)Xu, Xian, Wang, Schiele, and Akata]{attribute-proto}
Wenjia Xu, Yongqin Xian, Jiuniu Wang, Bernt Schiele, and Zeynep Akata.
\newblock Attribute prototype network for any-shot learning.
\newblock \emph{Int. J. Comput. Vis.}, 130\penalty0 (7):\penalty0 1735--1753, 2022.

\bibitem[Zhang et~al.(2020)Zhang, Cai, Lin, and Shen]{deepemd}
Chi Zhang, Yujun Cai, Guosheng Lin, and Chunhua Shen.
\newblock Deepemd: Few-shot image classification with differentiable earth mover's distance and structured classifiers.
\newblock In \emph{CVPR}, 2020.

\bibitem[Zhang et~al.(2023{\natexlab{a}})Zhang, Wu, Zhang, Wang, and Xie]{knowledge-guide}
Xiaoman Zhang, Chaoyi Wu, Ya Zhang, Yanfeng Wang, and Weidi Xie.
\newblock Knowledge-enhanced visual-language pre-training on chest radiology images, 2023{\natexlab{a}}.

\bibitem[Zhang et~al.(2023{\natexlab{b}})Zhang, Gao, Zhou, Wang, Qiao, Zhang, and Wang]{CITE}
Yunkun Zhang, Jin Gao, Mu Zhou, Xiaosong Wang, Yu Qiao, Shaoting Zhang, and Dequan Wang.
\newblock Text-guided foundation model adaptation for pathological image classification.
\newblock In \emph{MICCAI}, 2023{\natexlab{b}}.

\bibitem[Zhang et~al.(2023{\natexlab{c}})Zhang, Huang, Ma, Li, Luo, Xie, Qin, Luo, Li, Liu, Guo, and Zhang]{ram}
Youcai Zhang, Xinyu Huang, Jinyu Ma, Zhaoyang Li, Zhaochuan Luo, Yanchun Xie, Yuzhuo Qin, Tong Luo, Yaqian Li, Shilong Liu, Yandong Guo, and Lei Zhang.
\newblock Recognize anything: {A} strong image tagging model.
\newblock \emph{CoRR}, abs/2306.03514, 2023{\natexlab{c}}.

\bibitem[Zhou et~al.(2022)Zhou, Yang, Loy, and Liu]{coop}
Kaiyang Zhou, Jingkang Yang, Chen~Change Loy, and Ziwei Liu.
\newblock Learning to prompt for vision-language models.
\newblock \emph{Int. J. Comput. Vis.}, 130\penalty0 (9):\penalty0 2337--2348, 2022.

\end{thebibliography}
}

\end{document}